# Optimized Loss Functions for Object detection: A Case Study on Nighttime Vehicle Detection

Shang Jiang, Haoran Qin, Bingli Zhang*, Jieyu Zheng

School of Automotive and Transportation Engineering, Hefei University of Technology, Hefei, China

**Abstract:**

Loss functions is a crucial factor that affecting the detection precision in object detection task. In this paper, we optimize both two loss functions for classification and localization simultaneously. Firstly, by multiplying an IoU-based coefficient by the standard cross entropy loss in classification loss function, the correlation between localization and classification is established. Compared to the existing studies, in which the correlation is only applied to improve the localization accuracy for positive samples, this paper utilizes the correlation to obtain the really hard negative samples and aims to decrease the misclassified rate for negative samples. Besides, a novel localization loss named MIoU is proposed by incorporating a Mahalanobis distance between predicted box and target box, which eliminate the gradients inconsistency problem in the DIoU loss, further improving the localization accuracy. Finally, sufficient experiments for nighttime vehicle detection have been done on two datasets. Our results show than when train with the proposed loss functions, the detection performance can be outstandingly improved. The source code and trained models are available at https://github.com/therebellll/NegIoU-PosIoU-Miou.

## 1 Introduction

With the development of deep learning technology and computing power, CNN (convolutional neuro network) based object detection models are widely applied on intelligent driving or other industries. Most of those models can be divided into one-stage detection and multi-stage detection. The former is represented by YOLOs[1][2], SSD[3], RetinaNet[4] and EfficientDet[5], while the latter is represented by Faster-RCNN[6], Mask-RCNN[7] and Cascade RCNN[8]. Besides, a large amount of methods has been developed for improving detection accuracy or speed, with optimizing backbone[9][10], feature fusion[11], loss function[12]-[14], NMS (Non-Maximum Suppression)[15] etc. And this paper will further optimize the loss functions based on current state-of-art design.

The loss functions of object detection can be categorized as two sorts: the classification loss and the localization loss. The former is applied to train the classify head for determining the type of target object, and the latter is used to train another head for regressing a rectangular box to locate target object.

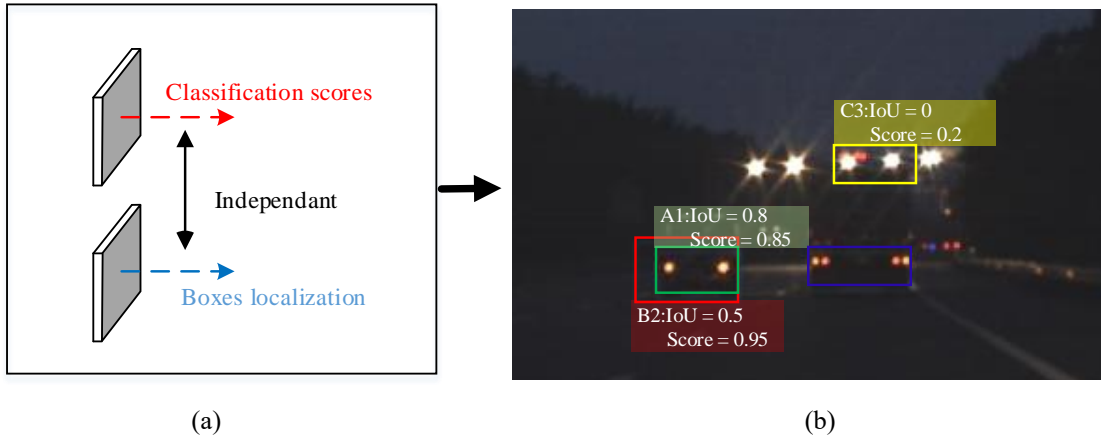

(a)　　　　　　　　　　　　　　　　(b)

*Fig 1 (a) Independent subnetworks for classification and localization; (b) Predicted box A1 with higher localization accuracy is suppressed by box B2 in NMS process; False detection C3 for negative sample whose IoU is zero.*

From the loss functions definition, we can notice that the heads for classification and localization are trained respectively with full independent loss functions. The classification scores are only determined by the corresponding feature map and classification head, regardless of the localization accuracy, which represented by the IoU (Intersection over Union) between predicted box and target box. Thereby, as mentioned in many existing studies, the predicted rectangular box that has higher IoU may has lower classification score, then could be suppressed by the following NMS process. As shown in the subfigure (b) of Fig 1, the box A1 (the yellow one) with higher IoU is suppressed by box B2 (the red one), although A1 has more accurate localization than B2. By adding IoU predicted head[16] or using IoU-based label[17], the IoU information from localization regression has been successfully combined to calculate the classification score of positive samples. The predicted boxes with higher IoU tend to have higher classification scores, and can be selected as the final output boxes after the subsequent NMS process. However, few studies have been done to explore the combination of IoU in negative samples detection. As shown in the subfigure (b) of Fig 1, in the nighttime vehicle detector, if the classification head is not design well, the paired two street lights may be misclassified as a vehicle, as the structure distribution of them is so similar to tail lights of vehicles, but with combining the IoU information from localization head, the paired two street lights could be detected as background easily as the corresponding IoU equals to 0.

Besides, as for the optimization of localization loss function, it was developed from $ln$-norm loss [3][4] into IoU loss[12] and GIoU loss[13] successively, and until so for, DIoU loss[14] has been developed and proved to be the best localization loss. Compared to the IoU and GIoU loss, a normalized distance between the predicted box and target box is incorporated to solve the slow convergence and inaccurate regression problems.

$$R_{DIoU} = \frac{\rho\left(\mathbf{b}, \mathbf{b}^{gt}\right)}{c^2} \qquad (1)$$

As shown in formular (1), $\rho$ denotes the actual centre distance between the predicted box and target box, while $c$ represents the diagonal length of the smallest enclosing box covering the two boxes. The aim of normalized distance loss $L_{DIoU}$ is prompting the centre of predicted box to get close to the centre of target box with the fastest speed. However, we can notice that with the decrease of centre distance $\rho$, the diagonal length $c$ reduces at the same time, leading the decrease gradient of $L_{DIoU}$ become inconsistent with the decrease gradient of $\rho$.

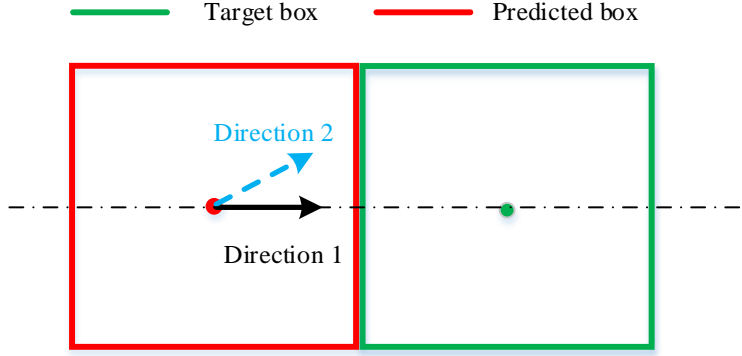

*Fig 2  Gradients inconsistency between normalized distance $L_{DIoU}$ and actual distance ρ, hurting the convergence speed.*

As shown in Fig 2, with the direction 1, represented by black solid arrow, the centre of predicted box can get close to the centre of target box in the fastest way, but the decrease gradient of loss $L_{DIoU}$ is in direction 2, shown as blue dot arrow. Thus, although the DIoU improved the convergence speed of the predicted box centre significantly, it is not be the best choice. By unifying the two gradients of loss $L_{DIoU}$ and centre distance ρ more reasonably, the convergence speed and whole localization accuracy can be further improved.

To address the above two issues, this paper optimizes the classification loss and localization loss respectively. Firstly, in order to establish the correlation between classification and localization subnetworks, the classification loss is changed to the production of CE (cross entropy) loss and an IoU-based coefficient. Specially, unlike the previous researches only combining IoU information with the positive sample for improving localization accuracy, this paper also applies IoU information to negative samples classification and decrease the false detection rate successfully. Besides, the function form of IoU-based coefficient is discussed and the corresponding parameters have been tuned for better performance. Secondly, on behalf of solving the gradients inconsistency between DIoU loss and fastest closing trajectory of predicted box centre, we propose a novel MIoU with incorporating Mahalanobis Distance. Finally, the optimized loss functions have been applied to a nighttime vehicle detection model. The baseline is a modified SSD detection network and the evaluation is carried out on two different nighttime vehicle datasets. The effectiveness of our proposed method is validated through experiments.

The main contributions of our work are as follows:

1) We propose a new classification loss function with combining IoU information to establish correlation between classification and localization. Unlike other researches only applying the correlation to improve the localization accuracy of positive samples, this paper successfully reduces the misclassified rate of negative samples at meantime. In the nighttime vehicle detection situation, the false detections caused by those environmental lights having similar structure with vehicle tail lights is reduced largely with the proposed classification loss function.

2) On the basis of DIoU loss, a more reasonable localization loss MIoU based on Mahalanobis distance has been proposed, eliminating the gradients inconsistency problem in DIoU loss. In the nighttime vehicle detection situation, the localization accuracy of vehicles has been further improved.

## 2 Related Work

### 2.1 Combination of Localization and Classification

IoU-Net proposes an additional subnetwork learning to predict the IoU between each detected bounding box and the target box in the baseline of Faster-RCNN [16]. During the inference time, all positive samples are ranked by the predicted IoUs instead of the classification scores and the anchor having the highest predicted IoU with target box will be selected to eliminate all other boxes having an overlap greater than a given threshold. Similarly, in the object segmentation application, a mask IoU subnetwork has been appended to the original Mask R-CNN network [18], and then during inference, the final mask confidence for NMS is calibrated by the production of predicted mask IoU and classification score. In the one-stage object detection, an IoU-aware object detector has been proposed based on the RetinaNet [19], by attaching an IoU head parallel with the classification and localization heads, and the predicted IoUs are multiplied by the classification scores as the final detection confidence for all positive samples. In the multiple anchor learning strategy [20], anchor bags are constructed, and by selecting the most representative anchors who have the best joint classification and localization confidence, the anchors with higher localization accuracy have been chosen. Instead of introducing an additional parallel head or anchors, other researchers assign different weights in the classification loss based on the IoUs between predicted boxes and target boxes, to increase the correlation between classification and localization. PISA [21] introduces an HLR algorithm based on IoU or classification score to rank the importance of samples and distribute different weights for each samples, and then introduce a new loss called classification-aware regression loss to jointly optimize both the classification and localization branches. Similarly, the IoU balanced classification loss [22] uses the predicted IoUs to modify the classification loss of positive samples directly, aiming to enhancing the attention on those positive samples with higher IoU during training process. GFL [17] introduces a classification vector where its value at target box category index refers to its corresponding IoU between predicted box and corresponding target box. By this configuration, the classification score and IoU are unified to a joint and single variable, which is applied to calculate the new classification loss during training time. We can notice that all the above methods only combine the information of IoU with the positive samples, aiming to enhance the localization accuracy. However, few researches have been done to explore the combination of IoU information with negative samples detection.

### 2.2 Evolution of Localization Loss Function

The $l_n$-norm loss [23] is introduced for box regression firstly, including the loss for centre points distance and the loss for box's height and width. With normalization method using height and width of predefined anchors, the $l_n$-norm loss could be applied to predict boxes with different size. The $l_n$-norm loss assumes that the object bounds to be four independent variables and regressed separately, resulting to less accurate localization. Thus, IoU loss function for bounding box prediction is introduced, which regresses the four bounds of predicted box as a whole unit [12]. Specially, as in the popular evaluation method COCO, IoU is an important metric for calculating average precision, the IoU loss function performs remarkable improvement on localization accuracy compared to the $l_n$-norm loss. But the IoU value will be zero if two objects do not overlap, and at this time it cannot reflect how far the two shapes are from each other. In the case of non-overlapping, the gradient will be zero and the IoU loss cannot be optimized. To address this issue, the GIoU loss [13] is proposed, by adding a penalty term based the smallest enclosing box covering the predicted and target boxes, and the penalty term will motivate the predicted box to move towards the target box. At present, researchers find that the GIoU loss still has slow convergence and inaccurate regression problems. The GIoU loss will totally degrade to IoU loss in full enclosing bounding boxes condition. And then the DIoU loss[14] for bounding box prediction is proposed, by incorporating the normalized distance between the predicted box and target box. However, as analysed in Section 1, the DIoU still has gradients inconsistent problem, which will be discussed and tackled in this paper.

## 2.3 Nighttime Vehicle detection

Nighttime vehicle detection has become one key task for advanced driver assistance systems (ADAS) and autonomous driving systems (ADS) in recent decades. About 30% of all vehicular accidents are caused by rear-end collisions that are one of the most fatal traffic accidents[24]. Compared to the vehicle detection in the daytime, nighttime detection seems to be more challenging, as the contrast between background and object, and the overall brightness are so low that some details of vehicles become unclear[25]. Early studies for nighttime vehicle detection are mainly based on vehicle lights detection and lights pairing technology. Specifically, based on Laplacian of Gaussian operator, a method named CensurE is introduced to detect the blobs in high speed, and then with SVM (support vector machines) classifier, those blobs belongs to vehicle lights are detected[26]. Besides, A nighttime image enhancement method is applied and the features extracted by CNN, HOG (histogram of oriented gradient), and LBP (local binary pattern) are fused to train the SVM classifier for vehicle blob detection[27]. With respect to the lights pairing, it is mainly based on prior knowledge, such as the relationship in shape and distance between two lamps. In order to improve the pairing accuracy, the SVM classifier is applied to the vehicle candidates obtained by lights pairing[28]. Unfortunately, the detection of vehicle lights is unstable as the vehicle lights is prone to be affected by the ambient lights like road lamps or traffic lights. To address this issue, at present, some CNN-based methods are also applied to nighttime vehicle detection. By extracting the high-level semantic features of vehicle itself, the detection performance is improved with comparing to only using vehicle lights information[29]. Besides, the correspondingly reflected lights in high-confident visual features fused for achieving further performance[30]. And GAN-based network is introduced to solve the vehicle detection problem during Day-to-Night transfer process[31]. However, in complex lighting conditions, the false detection rate caused by ambient lights is still high and the localization accuracy for remote vehicles is also need to be improved.

## 3 Proposed Method

### 3.1 Revisit to Presented Training Loss

In the anchor-based object detection tasks, the training loss is composed of two parts: the localization loss (loc) and classification loss (cls), and the overall objective loss is a weighted sum of them [3]:

$$L(c,l,p,g) = \frac{1}{N}\left(L_{cls}(c,p) + \alpha L_{loc}(l,g)\right) \tag{2}$$

Where, N is the number of positive samples, $\alpha$ is the weight term, $c$ and $p$ denote the predicted classes confidences and the label of target box respectively, $l$ and $g$ represent the localization vectors of predicted box and target box, consisting of central point coordinate $(cx, cy)$ and box's width and height $(w, h)$. Please note that in the actual anchor-based object detection model, the localization of predicted box is not regressed directed, and actually regress its offsets compared to the predefined anchor firstly. However, in order to illustrate clearly, we represent all the loss functions with absolute localization vector in this paper.

For each anchor, the classification loss is obtained with Softmax and Cross Entropy (CE):

$$L_{cls}^k(c, p) = -\log(\hat{c}^p), \text{ where } \hat{c}^p = \frac{\exp(c^p)}{\sum_p \exp(c^p)} \tag{3}$$

Besides, in order to eliminate the imbalance between positive and negative training examples, the hard negative mining is applied[3]. The negative examples with highest classification loss are selected and the ratio between negatives and positives is always set to 3:1.

With regard to the localization loss, the most advanced technology at present is DIoU loss, consisting of two parts: IoU loss part and the normalization centre distance part. For each anchor, the localization loss is shown as formular (4).

$$L_{loc}^k = 1 - \text{IoU} + \frac{\rho^2\left[\mathbf{b}(l^{cx}, l^{cy}), \mathbf{b}^{gt}(g^{cx}, g^{cy})\right]}{\lambda^2} \tag{4}$$

Where, IoU is the intersection over union between predicted box and target box, $\mathbf{b}$ and $\mathbf{b}^{gt}$ denote the central points of predicted box and target box respectively, and $\lambda$ represents the diagonal length of the smallest enclosing box covering the two boxes and can be calculated with the the localization vectors $l$ and $g$.

### 3.2 Classification Loss Optimization

As introduced in Section 1, by establishing correlation between localization and classification subnetworks, the localization accuracy of predicted box can be increased while the false detection rate of negative examples can be suppressed. Thus, this paper proposes a novel classification loss formular, with multiply an IoU-based coefficient $f(\text{IoU})$ by the standard CE loss, as shown in formular (5).

$$L_{cls}^k(c, p) = -f(\text{IoU})\log(\hat{c}^p) \tag{5}$$

The representation of IoU-based coefficient is inspired by the focal loss[4], which introduces a confidence-based scaling factor decaying to zero when confidence in the correct class increases. Similarly, we also introduce a positive tunable parameter $\gamma$ to achieve better performance. The specific representation of $f(\text{IoU})$ is shown in formular (6).

$$f(\text{IoU}) = \begin{cases} 1-(1-\text{IoU})^\gamma, & \text{positive anchor} \\ (1-\text{IoU})^\gamma, & \text{negative anchor} \end{cases} \tag{6}$$

We can notice that the expressions of IoU-based coefficient are different between positive and negative samples. For the positive samples, the coefficient $f(\text{IoU})$ is rising with the increasing of IoU, while for the negative samples, the variation tendency is opposite, as shown in Fig 3.

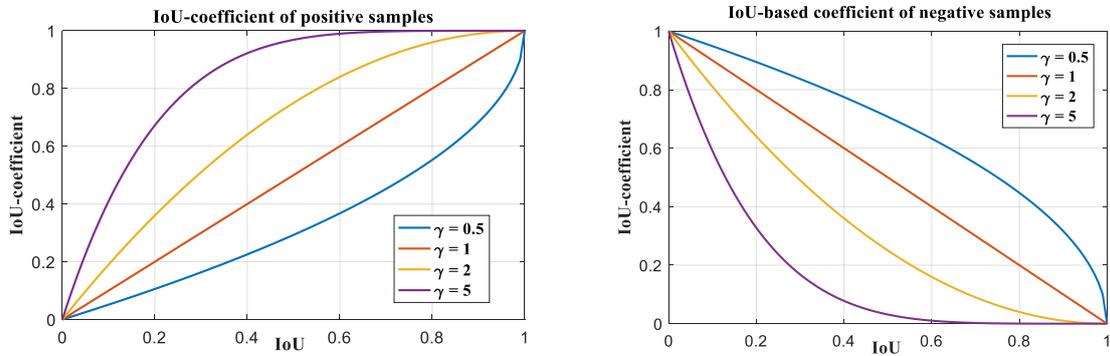

(a) (b)

*Fig 3 Variation of IoU-based coefficient with different IoU and $\gamma$: (a) positive samples; (b) negative samples*

Then, for the positive samples, those anchors with high IoU is up-weighted while those anchors with low IoU is down-weighted. For instance, with $\gamma = 2$, a positive sample with IoU = 0.95 would have 4 times larger loss than another positive sample with IoU = 0.52. As a result, the anchors with higher IoU are more likely to have larger CE loss, and can obtain larger gradients during training process, leading the model to learn high classification scores for them. On the contrary, the anchors with relatively lower IoU have smaller CE loss and smooth gradients, thus the corresponding classification scores can be suppressed after iterative training for some epochs. Then the predicted box with higher IoU and more accurate localization will have larger classification score, and will not be suppressed in the subsequent NMS process.

In order to illustrate the improvement for negative sample detection with formular (6), we put forward three misclassified negative samples at first, show in Fig 4, an enlarged view of nighttime vehicles in the distance.

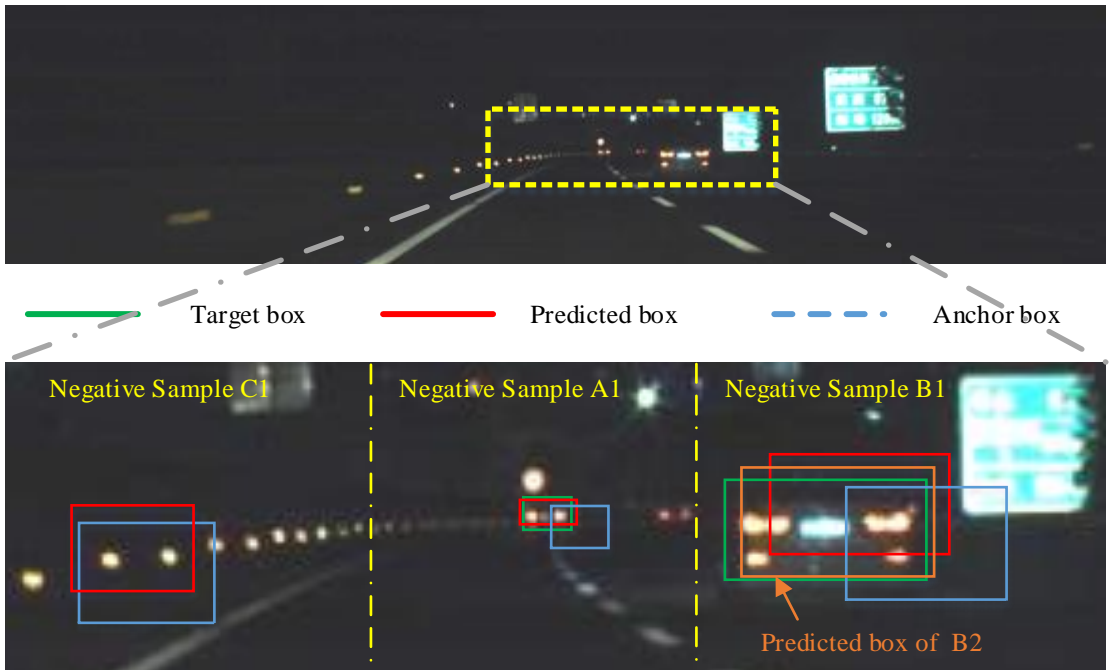

*Fig 4 Three misclassified negative samples (IoU between predefined anchor and target box is lower than 0.5) with different IoU between predicted box and target box. Green denotes the target box, red denotes the predicted box and blue denotes the predefined anchor box.*

Firstly, in the middle subfigure, A1 is judged as a negative sample as the IoU between its predefined anchor and the target box is lower than 0.5, but misclassified as a vehicle by the classification subnetwork. However, the predicted box of A1 calculated by localization subnetwork has a large IoU (0.8) with the target box. According to the COCO challenge[32], A1 should be determined to be true detection as its IoU is larger than the defined IoU threshold (0.5). By classifying A1 to a vehicle type, the mAP of detection model will be improved instead. Thus, no penalty or loss should be pay for the A1 classification. We decrease the standard CE loss to almost 0 by multiplying a very small coefficient (0.04) based on formular (6).

Secondly, in the right subfigure, negative sample B1 has a predicted box with an IoU equals to 0.45, and it should be determined to be false detection depending on the IoU threshold (0.5). However, as its IoU is close to the threshold, it has the possibility to be suppressed by adjacent true detection during the NMS process. As shown in the subfigure, B2 is a true positive prediction, and the IoU (0.7) between B1 and B2 is larger than the NMS suppression threshold (0.5). If the classification score of B2 is larger than B1, B1 will be suppressed by B2 and has no effect on the detection precision. Thus, we pay moderate penalty on this condition to push the model learn to classify B1 with a relatively low confidence when B1 is misclassified as a positive prediction.

Finally, in left subfigure, negative sample C1 has a predicted box with an IoU equals to 0. Unlike A1 or B1, C1 has no potential to be a true detection and no adjacent true detection could be found to suppress it. Thereby, negative samples like C1 should be defined as the real hard negative samples, which must be focused on and given large penalty during training stage. The IoU-based coefficient for the negative sample C1 is set to maximum to prompt the model learn to classify C1 as negative sample.

In conclusion, the design of IoU-based coefficient for negative samples can be divided into three range, shown in Fig 5 For the high IoU range samples, small coefficient is applied to reserve the true detections which are regarded as negative samples caused by unreasonable anchor setting; And then for the middle IoU range samples, moderate coefficient is applied to guarantee the classification confidence as a positive detection to be relatively low and then can be suppressed in NMS process; At last for the low IoU range samples, which should be defined as the real hard negative samples, large coefficient is used to ensure those samples can be classified as negative samples correctly.

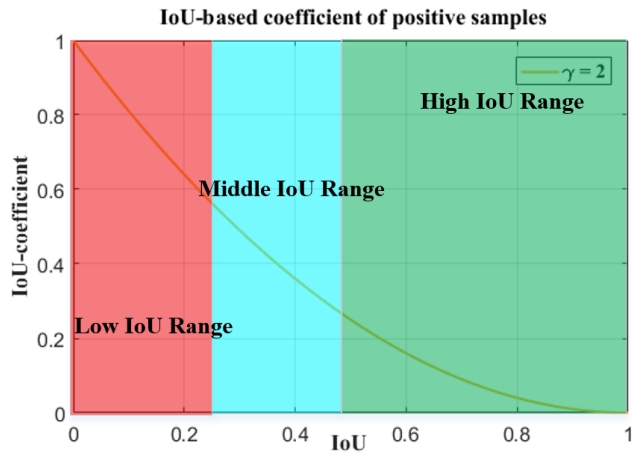

*Fig 5 Different configurations of IoU-based coefficient for three IoU ranges.*

Please note that the gradient calculation of IoU-based coefficient $f(\text{IoU})$ should be set to false in the back propagation. Because we do not hope the classification loss variable has any effect on the weight iteration of localization subnetwork, and by eliminating the gradient of $f(\text{IoU})$, the model can focus on the optimization of classification scores for both positive and negative samples.

### 3.3 Localization Loss Optimization

As introduced in Section 1, although the DIoU has improved the convergence speed and localization accuracy compared to other box regression loss functions, the gradients inconsistent between $L_{DIoU}$ and distance $\rho$ can still hurt its performance, as shown in Figxx. And in this chapter, we will model the gradient variation first, and then propose our method to address this issue.

As shown in Fig 6, a coordinate using the central point of target box is established, and in order to illustrate clearly, we suppose that the height and width of both target and predicted box are $a$.

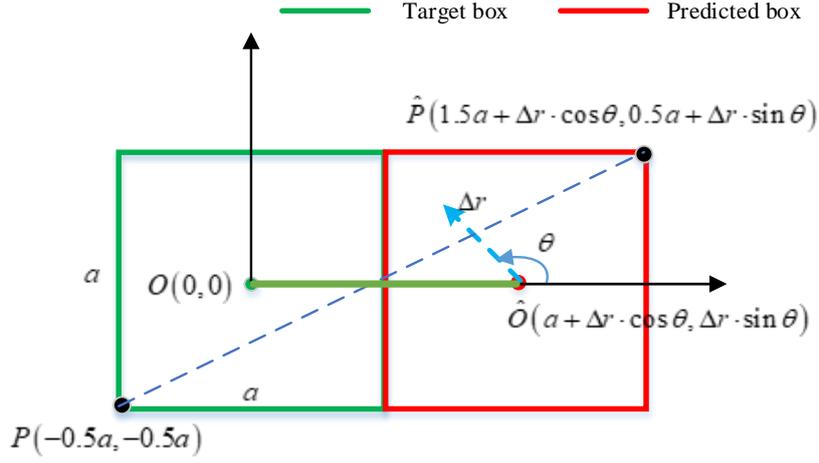

*Fig 6 Modelling of gradients inconsistency between DIou loss and fastest convergence.*

Besides, we define the moving vector of predicted box with distance $\Delta r$ and direction $\theta$, and then based on the definition of DIoU loss in formular (1), the normalized distance can be represented as:

$$R_{DIoU} = \frac{|P\hat{P}|^2}{|O\hat{O}|^2} = \frac{(a+\Delta r \cdot \cos\theta)^2 + (a+\Delta r \cdot \sin\theta)^2}{(2a+\Delta r \cdot \cos\theta)^2 + (2a+\Delta r \cdot \sin\theta)^2} \quad (7)$$

With setting a ten times length of $\Delta r$ to $a$, the variation of $R_{DIoU}$ with direction $\theta$ is represented in Fig 7. We can find that $R_{DIoU}$ will reach the minimum loss when $\theta$ equals to about 157°. But the optimal direction for minimizing distance between two boxes is horizontal by choosing $\theta$ with 180°.

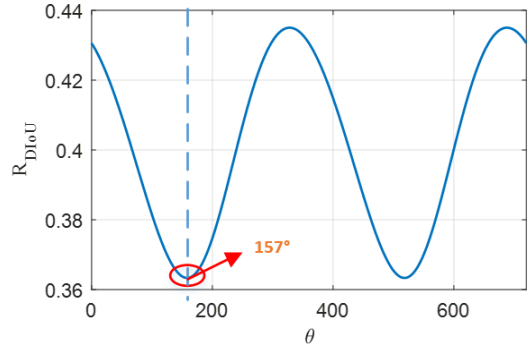

*Fig 7 Variation of DIoU loss with different moving direction, and the minimum loss is reached with $\theta$ equals to 157°.*

Thereby, inspired by the Mahalanobis distance, we propose a modified central points distance loss called MIoU, shown in formular (8).

$$R_{MIoU} = \left(\frac{l^{cx} - g^{cx}}{W}\right)^2 + \left(\frac{l^{cy} - g^{cy}}{H}\right)^2, \text{Where} \quad W = \frac{l^w + g^w}{2}, H = \frac{l^h + g^h}{2} \quad (8)$$

The difference of central points in width and height dimensions are normalized by introducing the sum of two boxes' width and height respectively, and $R_{MIoU}$ is actually the Mahalanobis distance of the two central points. Compare to the standard DIoU, the MIoU eliminate the scaling difference between height and weight dimension, the central point distance can be regressed with the same speed. In addition, we choose width and height as denominators, which are independent from the central points distance. The reduction of the central point distance will not change the variation of width or height, thus the gradients inconsistent problem existing in DIoU is also eliminated. In Fig 6, the gradients of MIoU loss and distance of $O\hat{O}$ will be identical if we replace DIoU with MIoU.

Similarly, in order to prevent the using the increase of W or H to minimize the $R_{MIoU}$, leading the decrease of convergence speed for central points, the gradient calculation of W and H are set to false in the back propagation.

## 4  Experiments

To validate our optimized loss functions, we perform sufficient evaluation experiments on detection of vehicle at nighttime. The specific process is introduced in the following paragraphs.

### 4.1 Dataset and evaluation metrics

We evaluate our proposed method on Dataset1 firstly, which is collected by ourselves and include the nighttime vehicles on the highway condition. The Dataset1 contain 11924 vehicles from 5465 images with 1280×720 resolution. We divide the whole dataset into training set, test set and validation set according to the ratio 3:1:1 in time domain. In addition, in order to validate our method more generally, we select the nighttime images of vehicle from Boxy dataset[32] and constructed another dataset called Dataset2. The Dataset2 contain 31705 vehicles from 6044 images with 1232×1028 resolution. Compared to Dataset1, the vehicle distribution in Dataset2 is denser, and the urban road condition is also included in Dataset2.

We apply the COCO-style AP metrics in this work, consisting of AP (averaged AP at IoUs from 0.5 to 0.95 with an interval of 0.05), AP75(AP at IoU threshold 0.75), $AP_S$(AP for objects with width pixels less than 32), $AP_M$(AP for detected objects with width pixels between 32 and 96), $AP_L$(AP for detected objects with width pixels larger than 96). As we find that the vehicle distribution in both two datasets are mainly focus on the objects with pixels less than 96, thus in our following experiments, we will not analysis the $AP_L$ metric.

### 4.2 Base network

We choose the popular one-stage detector SSD[3] as our base network, and in order to achieve better performance in nighttime vehicle detection task, we make some modifications on the original SSD based on the Pytorch platform.

1) As the images in those two datasets are not square, the input image size such as [512, 512] is not suitable in our application, thus we resize the input size of network to the half of the original images. The resized input sizes will be [640, 360] and [616, 514] for Dataset1 and Dataset2 respectively.

2) We replace the VGG16 with ResNet-50, as the latter one has been proved to have strong performance on both detection accuracy and execution speed.

3) We apply K-means clustering[2] approach on the datasets and obtain the most suitable anchors for this detection task. In particular, we find that the vehicles are mainly distributed in relatively lower pixels range. By considering the trade-off between speed and accuracy, we finally apply one anchor with pixels [57, 48] to the output of conv3_x layer, and two anchors with pixels [87, 71] and [136 135] to conv4_x layer for the Dataset1. And for Dataset2, the corresponding anchors are [19,17] in conv3_x layer, and [71,60] in conv4_x layer[34].

4) The detection heads for both classification and localization are both replaced by a 3×3 convolutional followed by a 1×1 convolutional layer to improve the nonlinear fitting capability.

Finally, two 1080Ti GPUs are applied in the training process. Each minibatch include 8 images, and all the models in this work are trained for 30 epochs with SGD (stochastic gradient descent) algorithm. Weight decay of 0.0005 and momentum of 0.9 are used, and the learning rate will be a half after every 5 epochs.

### 4.3 Main Results on Dataset1

As shown in Table 1, we choose the standard loss functions as baseline, in which the classification loss has no correlation with localization accuracy and the loss function for regression box is $l_1$-norm loss. Firstly, we compare the detection precision by using optimized classification loss function. Pos-IoU is the case that the IoU-based coefficient only applied to calculate the classification loss of positive samples while Neg-IoU is opposite. We can notice that the Pos-IoU can impove 1.7% AP and 3.9% AP75 while Neg-IoU can improve 4% AP and 6.8% AP75, which means that the combination of IoU information for negative samples is more effective than its application to positive samples. In addition, by applying Pos-IoU and Neg-IoU together, a higher performance has been reached with improving 4.7% AP and 7.8% AP75. Secondly, as for the optimization of localization loss function, the results show that the DIoU improve 1.2% AP and 3.9% $AP_{75}$, but incur a slightly decrease in the $AP_M$ metric. However, the MIoU get both improvement in all metrics, with 3.8% $AP_S$ and 0.6% $AP_M$ respectively. Finally, we optimize both localization loss and classification loss together, and then obtain the highest performance with improving 6.5% AP and 11.3% AP75. The main results demonstrate the powerful capability of our methods for detection precision improvement.

Table 1: Quantitative comparison with different loss optimization methods on Dataset1 with α = 1.0 and γ = 2.0.

| Optimized approaches | Pos-IoU | Neg-IoU | DIoU | MIoU | AP | $AP_{75}$ | $AP_S$ | $AP_M$ |
|---|---|---|---|---|---|---|---|---|
| Baseline | | | | | 38.1 | 31.6 | 24.7 | 49.6 |
| Classification loss | √ | | | | 39.3 | 34.7 | 25.1 | 49.3 |
| | | √ | | | 42.1 | 38.4 | 31.4 | 50.6 |
| | √ | √ | | | **42.9** | **38.3** | 30.5 | 50.8 |
| Localization loss | | | √ | | 39.3 | 36.5 | 29.4 | 48.7 |
| | | | | √ | 40.8 | 39.2 | 28.5 | 50.3 |
| Both loss functions | √ | √ | | √ | **44.6** | **42.9** | 30.3 | 51.4 |

Fig 6 is the specific visualization comparison results between baseline and our method. It is obvious that by applying the optimized loss functions, the false detection samples caused by surrounding illuminations have been eliminated. Besides, the localization accuracy for positive samples is also improved remarkably.

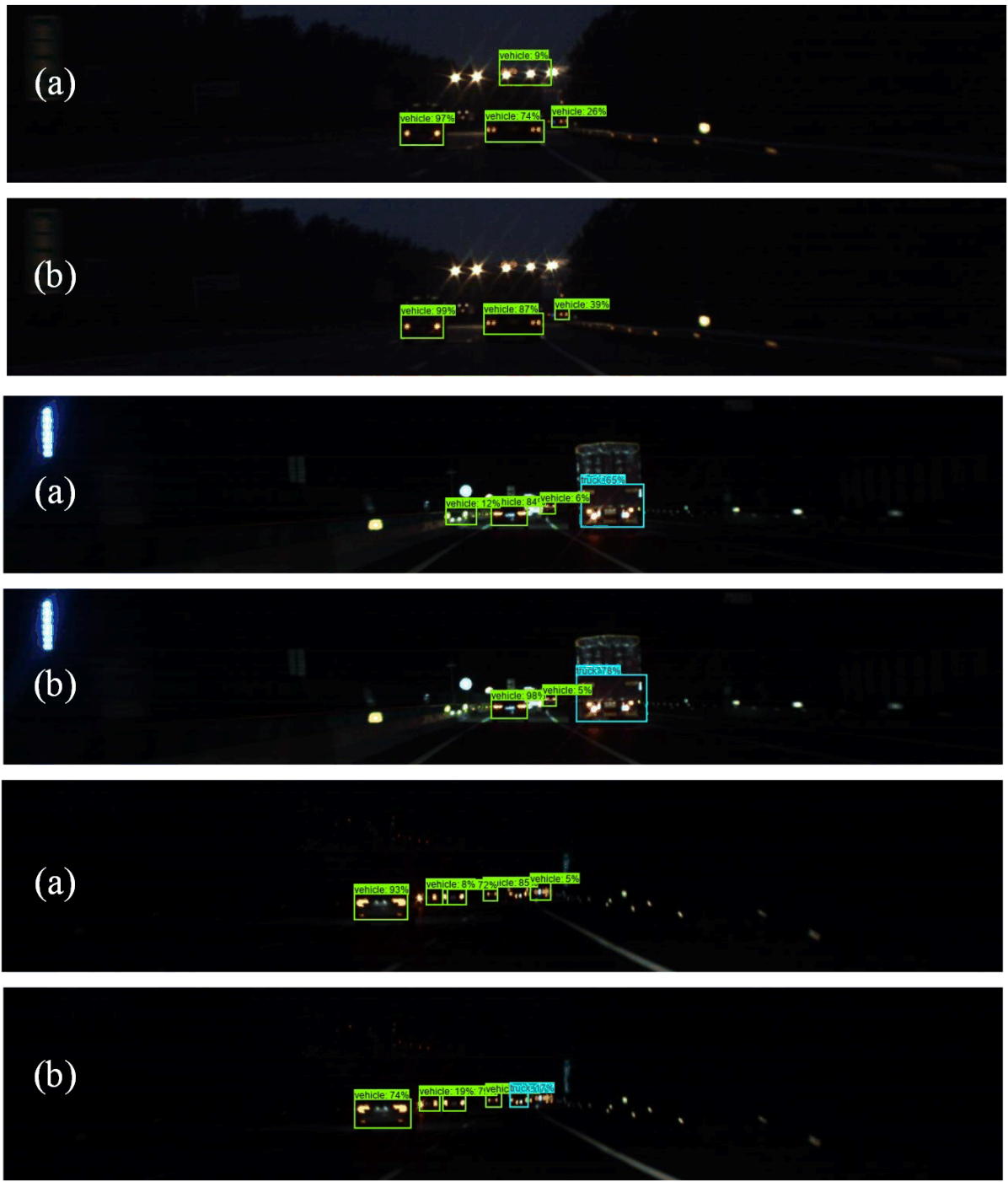

*Figure 6: Visualization results. (a): Baseline; (b): Our proposed method.*

### 4.4 Ablation Study on Dataset1

In order to find the optimal α (weight term between classification loss and localization loss) and γ (parameter of IoU-based coefficient in formular (6)), to achieve the highest performance based on the proposed solution including both two loss function optimization, we proceed an ablation study on Dataset1. Firstly, we fix α to 1.0 and try to find the range of optimal γ. As shown in Table 2, we find that AP and $AP_{50}$ both reach the maximum when $\gamma = 1.5$, thus we preliminary locate the optimal γ should be in the range between [1.0, 2.0].

Table 2: Quantitative comparison with different γ after incorporating both loss functions optimization with α = 1.0

| γ | AP | $AP_{50}$ |
|---|---|---|
| 0.5 | 43.4 | 41.6 |
| 1.0 | 44.1 | 81.6 |
| 1.5 | **45.4** | **83.0** |
| 2.0 | 44.6 | 80.3 |
| 2.5 | 43.3 | 79.6 |
| 5.0 | 38.5 | 73.7 |

And then, we give a more refined division to γ between range [1.0, 2.0], and at the same time, we change weight term α between range [0.8, 1.2] with step equals to 0.1. As shown in Table 3, when α is 1.0 and γ is 1.75, our proposed method reaches the maximum precision with improving 7.6% AP comparing to the baseline. The corresponding improvement in AP75, $AP_S$ and $AP_M$ are 42.9%, 30.1% and 52.2% respectively.

Table 3: Quantitative comparison with different γ and α after incorporating both loss functions optimization.

| α \ γ | 1.0 | 1.25 | 1.5 | 1.75 | 2.0 |
|---|---|---|---|---|---|
| 0.8 | 42.1 | 44.1 | 44.2 | 45.1 | 44.2 |
| 0.9 | 44.2 | 45.2 | 45.4 | 44.8 | 45.2 |
| 1.0 | 44.1 | 45.1 | 45.4 | **45.7** | 44.6 |
| 1.1 | 44.3 | 45.0 | 44.6 | 44.0 | 44.0 |
| 1.2 | 41.5 | 45.4 | 44.0 | 44.0 | 44.2 |

## 4.5 General Validation in Dataset2

In order to prove the generality of our proposed method, we also carry out similar experiments on Dataset2, whose vehicle distribution is denser and the corresponding environment is more complex. As shown in Table 4, by applying proposed classification loss and localization loss separately, all the precision metrics have been improved with different degrees. In particular, Neg-IoU lead the maximum contribution to precision improvement, by increasing 2.3% AP and 3.9%% AP75. And by combining all optimized loss function together, the detection precision is further improved to 45.9% AP and 44.8% AP75 at last.

Table 4: Quantitative comparison with different loss optimization methods on Dataset2

| Optimized approaches | Pos-IoU | Neg-IoU | DIoU | MIoU | AP | $AP_{75}$ | $AP_S$ | $AP_M$ |
|---|---|---|---|---|---|---|---|---|
| Baseline | | | | | 42.9 | 39.3 | 33.5 | 57.9 |
| Classification loss | √ | | | | 43.4 | 40.1 | 34.0 | 58.0 |
| | | √ | | | **45.2** | **43.2** | 33.8 | 59.3 |
| | √ | √ | | | 45.5 | 43.8 | 34.2 | 60.3 |
| Localization loss | | | √ | | 43.3 | 40.6 | 34.3 | 58.1 |
| | | | | √ | 43.4 | 40.2 | 34.2 | 58.4 |
| Both loss functions | √ | √ | | √ | **45.9** | **44.8** | 34.4 | 60.6 |

# 5  Conclusions

In this work, we optimized the classification and localization loss functions respectively. The main idea of classification loss function optimization is to combining localization information to the classification results, thus more accurate classification scores for both positive and negative samples are obtained. As for the optimization of localization loss function, we proposed the MIoU loss to further improve the localization accuracy on the basis of DIoU. Both two kind of loss functions optimization are applied to nighttime vehicle detection and the results show that the proposed loss functions is effective for detection precision improvement. Specially, we find that with combing the IoU information to the negative sample detection, the false detection rate for negative samples which have the similar structure with vehicle tail lights, can be decreased largely in complex light condition, leading 4.7% AP and 7.8% AP75 improvement respectively. We consider to apply this method to nighttime pedestrian detection, in which false detection of negative samples is the main reason hurting average detection precision.

## Acknowledgments

This research was supported by Anhui Intelligent Automobile Engineering Laboratory, and the corresponding grant number is PA2018AFGS0026.